# Harnessing Self-Supervised Deep Learning and Geostationary Remote Sensing for Advancing Wildfire and Associated Air Quality Monitoring: Improved Smoke and Fire Front Masking using GOES and TEMPO Radiance Data

Nicholas LaHaye
*Spatial Informatics Group, LLC.*
Pleasanton, CA, USA
nlahaye@sig-gis.com*

Thilanka Munashinge
University at Albany
*Albany, NY, USA*
tmunasinghe@albany.edu

Hugo Lee
*Jet Propulsion Laboratory*
California Institute of Technology
*Pasadena, CA, USA*
huikyo.lee@jpl.nasa.gov

Xiaohua Pan
*ADNET systems, INC., Bethesda, MD, USA,*
*NASA Goddard Space Flight Center*
*Greenbelt, MD,*
xiaohua.pan@nasa.gov

Gonzalo Gonzalez Abad
*Center for Astrophysics*
Harvard & Smithsonian
*Cambridge, MA, USA*
ggonzalezabad@cfa.harvard.edu

Hazem Mahmoud
*ASDC, ADNET,*
*NASA Langley Research Center*
Hampton, VA, USA
hazem.mahmoud@nasa.gov

Jennifer Wei
*NASA Goddard Space Flight Center*
*Greenbelt, MD,*
jennifer.wei@nasa.gov

**This work demonstrates the possibilities for improving wildfire and air quality management in the western United States by leveraging the unprecedented hourly data from NASA's TEMPO satellite mission and advances in self-supervised deep learning. Here we demonstrate the efficacy of deep learning for mapping the near real-time hourly spread of wildfire fronts and smoke plumes using an innovative self-supervised deep learning-system: successfully distinguishing smoke plumes from clouds using GOES-18 and TEMPO data, strong agreement across the smoke and fire masks generated from different sensing modalities as well as significant improvement over operational products for the same cases.**

## I. Introduction

Wildfires in the western U.S. have alarmingly increased and become more widespread after 2000 [1][2][3]. However, monitoring and predicting wildfire-induced air pollutants remain challenging. Wildfire smoke plumes undergo complex chemical transformations, especially in the first few hours after emission, and the evolution of wildfire-induced chemical compositions varies rapidly, both near-source and downwind [4]. The successful launch of the Tropospheric Emissions: Monitoring of Pollution (TEMPO) mission has brought a breakthrough. TEMPO can track these rapid changes in atmospheric chemical compounds released from wildfire combustion, including directly emitted nitrogen dioxide (NO2) and volatile organic compounds (VOCs) (such as formaldehyde, i.e., HCHO), and current and upcoming products measuring secondary pollutants like ozone (O3), almost near real-time (NRT) at an hourly frequency currently starting from August 2, 2023 (https://tempo.si.edu/index.html). Being the first space-based instrument to monitor air pollution (GOES with aerosols and TEMPO with trace gasses) hourly across Greater North America, TEMPO surpasses the limitations of Low Earth Orbit (LEO) satellites, like the TROPOspheric Monitoring Instrument (TROPOMI) which can observe a specific location only once per day. TEMPO data also helps fill gaps in ground-based monitor networks in North America at a higher spatial resolution of 2.0 km × 4.7 km2 at the center of the field of regard compared to 5.5 × 3.5 km2 for TROPOMI [5]. Accordingly, in this work, we harness TEMPO data's unprecedented advantages and self-supervised deep learning to take a step towards transforming wildfire and air quality (AQ) management in the western U.S enabling mapping of the NRT hourly spreads of wildfire fronts and wildfire-induced smoke plume masks for TEMPO by utilizing an innovative deep learning (DL) system called Segmentation, Instance Tracking, and data Fusion Using multi-SEnsor imagery (SIT-FUSE).

The products developed here directly add value to the existing TEMPO data for wildfire study and application, which meet the urgent need to improve smoke plume detection in TEMPO data. They are particularly valuable for enhancing disaster response, allowing agencies like the Federal Emergency Management Agency (FEMA) to issue timely and accurate wildfire watches and air quality warnings. This proactive approach will help protect lives and property in wildfire-prone areas of North America, especially in vulnerable communities. Here we demonstrate that using the Park Fire as a case study.

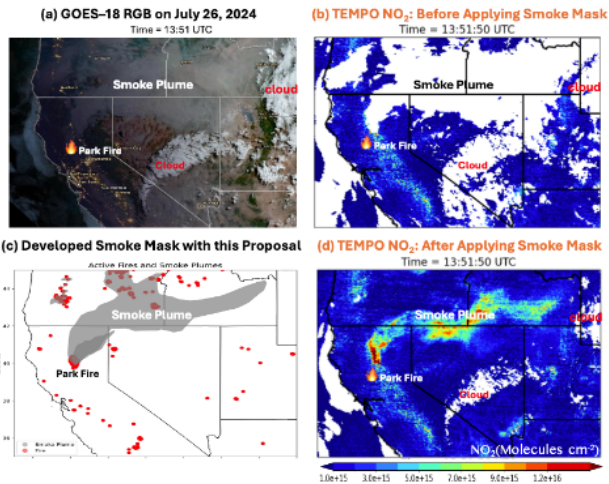

*Figure 1: The RGB image from GOES-18 on July 26, 2024, 14:51:50 UTC shows thick smoke plume during Park Fire in CA; (b) TEMPO L3 NO2 troposphere vertical column is filtered out in smoke plume (Units: Molecules cm-2) before applying smoke mask developed in this study (c), but (d) restoring values after applying the smoke mask.*

Figure 1 illustrates the need of developing smoke masks with a July 26, 2024, case study during the Park Fire. Most smoke plumes and clouds in GOES-18 red-green-blue (RGB) composite imagery are distinguishable with our eyes (Figure 1a). However, in the current TEMPO data, smoke plumes are misclassified as clouds often with cloud fractions exceeding 0.2, resulting in filtering retrievals of NO2 tropospheric vertical column downwind of the smoke plumes (Figure 1b), as per the TEMPO's L2 and L3 Trace Gas and Cloud user guide [6], i.e., maintaining good quality NO2 data by removing data where the cloud fraction exceeds 0.2. To address this issue, we propose to develop the smoke plume masks for TEMPO, which resemble the mockup in Figure 1c, to restore missing data in wildfire smoke plumes (Figure 1d).

## II. Methods

SIT-FUSE utilizes self-supervised representation learning and segmentation for challenging domains with limited or no label sets. This DL framework is hierarchical, allowing different levels of specificity based on dataset inputs and user needs. Once self-supervised segmentation maps are generated for the training set, a small subset of scenes is used to assign context. This can be done semi-manually via a Geographic Information System (GIS) or automatically if pre-existing label sets are available. This process ensures the subject matter experts are still in the loop while minimizing the time required to work on pre-analysis tasks. Information from the spatial distribution of self-supervised segmentation labels and shape characteristics serves as input for the temporal tracking element of this task. This DL framework has been successfully tested on varying spatial, spectral, and temporal resolutions, demonstrating its reliability in real-world scenarios. Capabilities have been successfully demonstrated on high and low spatial resolution data, multispectral to hyperspectral data, and polar-orbiting, airborne, and geostationary data for segmentation, characterization, and tracking use-cases including wildfires, smoke, algal blooms, and oil palm plantations in the Amazon [7,8,9].

SIT-FUSE is developed to be a generic framework allowing various kinds of encoders and foundation models that leverage self-supervised representation learning, including the simplest of the options, Deep Belief Networks (DBNs) trained using contrastive divergence, which were used here. For all of these experiments, we used DBNs with 2-3 layers. DBNs were selected here because previous work and experiments done have demonstrated that they produce reasonable results and the parameter space is much smaller than the other models available. We have done extensive validation on the use of DBNs from both the perspective of structural understanding and downstream task performance, as well as resource consumption assessment, for the large set of single-instrument and fusion datasets [8]. In short, 2-3 layer DBNs provide a relatively compact model (~2 million parameters vs ~100 million - 10 billion parameters) with representational capabilities that meet our needs, and have demonstrated out-sized representational capabilities in other experiments relative to much larger models [10]. We are currently evaluating encoder complexity in relation to segmentation performance and geographic coverage, to optimally operationalize this approach for operational global production, which will be discussed further in later manuscripts.

For the self-supervised segmentation, we leverage Deep Clustering techniques, specifically Information Invariant Clustering. To mimic the hierarchical nature of traditional agglomerative clustering output, we have set up hierarchical deep clustering layers. Here, the output heads are set up in a tree structure where each sub-tree is only trained on label samples classified as belonging to their parent label sets. Each layer receives only the output of the encoder, but as the tree is built from the top down, each neural network in a child node position on the tree only sees samples assigned labels associated with its parent and ancestor node(s). In this way, we can create scene segmentation at varying levels of specificity and explore the connections between each level. To our knowledge, this is the first study/ software system that leverages IIC layers, or even deep clustering layers, in this fashion.

For both context assignment and validation, subject matter experts labeled areas of high-certainty smoke, fire, and associated backgrounds. All areas that labellers were uncertain about, remained without labels. This labeling process was done by generating polygons over the remote sensing imagery. Because scenes can have overlapping classes (i.e. fire and smoke are contained in the same pixel), but also have areas distinct to a single class, a separate background class label set was generated for fire and smoke. Unlike labeling for supervised learning, this approach does not require all training samples to be labeled, which is relevant for problems with high uncertainty in boundary cases, like the segmentation of fire fronts and smoke plumes. The labeling of only the high certainty class areas allows us to capture and compare against segmentation structure, and provide ample samples for context assignment. Although this labeling minimizes the pre-analysis labor required from subject matter experts, it still keeps experts in the loop (a crucial piece for science-related ML tasks). Relative to the number of scenes labeled here, supervised and semi-supervised tasks require 100x more labeled samples or more. Also, because they are learning the mapping between the labels and the input datasets directly, they require much more complete label sets (i.e. uncertain areas must be labeled background or foreground, potentially leading to systematic over-segmentation or under-segmentation).

There is no direct ground truth here - meaning comparisons are done against pre-existing detections as well as hand labels generated over only the high certainty areas of smoke plumes and fire fronts - and we know there are other areas of these objects that are not included in the labels. Therefore, for validation purposes, recall, precision, and their collective summary via F1-scoring are too harsh of evaluators here. We evaluated some previously published versions of precision and recall that apply fuzzy logic, but ultimately landed on the structural similarity index (SSIM) to evaluate performance across the various dataset pairs [11,12]. This is a fairly common problem within the remote sensing domain and one we aim to help solve with the collective incorporation of self-supervised learning, subject matter expert domain knowledge, and large amounts of data [13,14].

### III. Results

Figure 2 demonstrates that our proposed "Wildfire Mapping System" can successfully distinguish between clouds and smoke and identify wildfire fronts. It shows the procedures of generating smoke masks and fire front on July 26, 2024, at 14:51 UTC, during the Park fire in CA, with GOES L1B radiance data as inputs (Figure 2a), followed by the subsequent intermediate outputs (Figure 2b and 2c), final wildfire fronts in red and smoke masks in blue (Figure 2d), and the TEMPO L3 NO2 troposphere vertical column data filtered by smoke masks (Figure 2e). Figure 3a-c depicts the smoke masks derived directly from TEMPO radiances for the scenes that temporally align most closely with the GOES scene from Figure 2b-d, respectively.

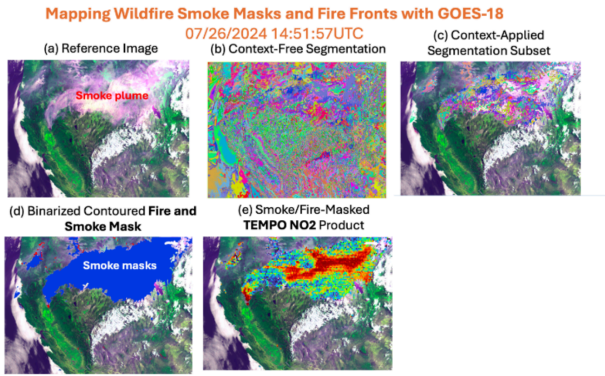

*Figure 2: Demonstration of successful GOES-18-based smoke and wildfire front masks for the Park fire on July 26, 2024, at 14:51 UTC. (a) Reference RGB from GOES-18, (b) and (c) subsequent intermediate outputs, (d) final wildfire fronts in red and smoke masks in blue, and (e) TEMPO L3 NO2 troposphere vertical column data over fire smoke masks.*

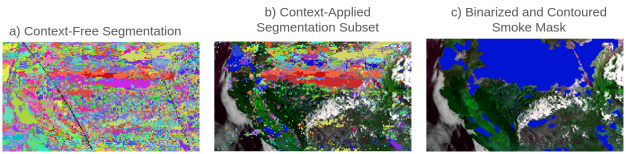

*Figure 3: Demonstration of successful TEMPO-based smoke masking for the Park fire on July 26, 2024, using scenes closest to the GOES RGB taken at 14:51 UTC. (a) and (b) subsequent intermediate outputs, overlaid on the RGB from Figure 2a, and (c) finalized smoke mask overlaid on the RGB in blue.*

Our validation set consists of 5+ scenes from 5 separate days the models did not see during training, consisting of > 7million pixels. When these were compared to operational products from VIIRS and GOES instruments, in cases during the Park fire where the operational products perform well, we attain an SSIM of 0.86 for smoke and 0.71 for fire. When compared against the hand labeled high-certainty smoke and fire pixels, we attain an SSIM of 0.83 for smoke and 0.7 for fire. There are also plenty of cases where operational products have misclassified the thick smoke plume as cloud and provided no retrieval of atmospheric compositions, such as NO2 and aerosols [14,15]. Both clouds and thick smoke plumes exhibit similar characteristics in satellite observations, e.g., high reflectivity and opacity. Consequently, the algorithm will misclassify the thick smoke as clouds.

## IV. Discussion

Here, each L1B dataset is being leveraged in a separate pipeline to take advantage of varying spatial, spectral, and temporal resolutions. The smoke and fire masks generated from each pipeline can be aggregated to create a combined sub-hourly smoke and fire progression mapping dataset. To combine the data from separate data streams into a single collective smoke and fire mask, we plan combining probabilistic label assignments from the ML models and the agreement between the spatiotemporally collocated smoke and fire masks in a way that allows for certainties in the range of [0.0,1.0], instead of strict binary masking. These kinds of datasets can easily be integrated into Earth System Digital Twins and modeling platforms like the Jet Propulsion Laboratory's Fire Alarm and Pyrecast. With these progression maps, and tools mentioned above, not only can the direct progression be better studied, and downstream retrieval coverage improved, but there are also more complete and granular datasets that can be leveraged for short term fire and smoke forecasting. Lastly, this data and forecasting capability can be incorporated into impact assessments and future fire, smoke damage, and air quality risk evaluations.

Another important piece is the intermediate context-applied segmentation subset (shown in figures 2c and 3b. This product distinguishes multiple subclasses within each binarized class identified. These subclasses alone provide a co-exploration space for subject matter experts to evaluate what parameterizations or features are being correlated to create these subclasses. Paired further with advanced deep-learning-based co-exploration tools, we believe this data provides an even richer analysis domain. Current research is being done to develop these tools.

## V. Acknowledgments

The authors would like to thank NASA, LaRC, The Atmospheric Science Data Center, GSFC, Center for Astrophysics, JPL, ADNet Systems, and The Spatial Informatics Group, LLC., for supporting this research. The authors would also like to thank the anonymous reviewers for taking the time to read this paper and provide valuable feedback.